\documentclass[letterpaper, 10 pt, conference]{ieeeconf}
\pdfoutput=1
\IEEEoverridecommandlockouts
\usepackage{url}
\usepackage{breakurl}

\title{\LARGE \bf
BinaryCoP: Binary Neural Network-based COVID-19 Face-Mask Wear and Positioning Predictor on Edge Devices}

\author{Nael Fasfous\textsuperscript{1*}, Manoj-Rohit Vemparala\textsuperscript{2*}, Alexander Frickenstein\textsuperscript{2*}, Lukas Frickenstein\textsuperscript{1}, Walter Stechele\textsuperscript{1} \\
$^{1}$ Technical University of Munich (\texttt{\small<first\_name>.<last\_name>@tum.de})\\ ~ $^{2}$ BMW Group (\texttt{\small<first\_name>.<last\_name>@bmw.de})
\thanks{$^{*}$ Equally contributed}
}

\usepackage{caption}
\usepackage{graphicx}
\usepackage{amsmath,amssymb} 
\usepackage{color}
\usepackage{listings}
\usepackage{diagbox}
\usepackage{subfig}
\usepackage{siunitx}
\usepackage{changes}
\usepackage{hhline}
\usepackage{makecell}
\usepackage{color, colortbl}
\usepackage{xcolor}
\usepackage{collcell}
\usepackage{multirow}
\usepackage{booktabs}
\usepackage{paralist}
\usepackage{pgfplots}
\usepackage{pgfplotstable}
\usepackage{filecontents}
\usepackage{tikz}
\usepackage{ifthen}
\usetikzlibrary{calc}
\usetikzlibrary{matrix}
\usetikzlibrary{patterns}

\usepackage{array}
\newcommand{\PreserveBackslash}[1]{\let\temp=\\#1\let\\=\temp}
\newcolumntype{C}[1]{>{\PreserveBackslash\centering}p{#1}}

\newcommand{\etal}{et al.~}
\newcommand{\ie}{\textit{i}.\textit{e}.~}

\newcommand{\bincop}{BinaryCoP}

\def\colorModel{hsb} %You can use rgb or hsb

\newcommand\ColCell[1]{
  \pgfmathparse{#1<50?1:0}  %Threshold for changing the font color into the cells
    \ifnum\pgfmathresult=0\relax\color{white}\fi
  \pgfmathsetmacro\compA{0}      %Component R or H
  \pgfmathsetmacro\compB{#1/100} %Component G or S
  \pgfmathsetmacro\compC{1}      %Component B or B
  \edef\x{\noexpand\centering\noexpand\cellcolor[\colorModel]{\compA,\compB,\compC}}\x #1
  } 
\newcolumntype{E}{>{\collectcell\ColCell}m{0.4cm}<{\endcollectcell}}  %Cell width

\begin{document}

\maketitle
\thispagestyle{empty}
\pagestyle{empty}

%%%%%%%%%%%%%%%%%%%%%%%%%%%%%%%%%%%%%%%%%%%%%%%%%%%%%%%%%%%%%%%%%%%%%%%%%%%%%%%%
%      __   __  ___  __        __  ___ 
% /\  |__) /__`  |  |__)  /\  /  `  |  
%/~~\ |__) .__/  |  |  \ /~~\ \__,  | 
%%%%%%%%%%%%%%%%%%%%%%%%%%%%%%%%%%%%%%%%%%%%%%%%%%%%%%%%%%%%%%%%%%%%%%%%%%%%%%%%
\begin{abstract}
Face masks have long been used in many areas of everyday life to protect against the inhalation of hazardous fumes and particles. They also offer an effective solution in healthcare for bi-directional protection against air-borne diseases. Wearing and positioning the mask correctly is essential for its function. Convolutional neural networks (CNNs) offer an excellent solution for face recognition and classification of correct mask wearing and positioning. In the context of the ongoing COVID-19 pandemic, such algorithms can be used at entrances to corporate buildings, airports, shopping areas, and other indoor locations, to mitigate the spread of the virus. These application scenarios impose major challenges to the underlying compute platform. The inference hardware must be cheap, small and energy efficient, while providing sufficient memory and compute power to execute accurate CNNs at a reasonably low latency. To maintain data privacy of the public, all processing must remain on the edge-device, without any communication with cloud servers. To address these challenges, we present \bincop{}, a low-power binary neural network classifier for correct facial-mask wear and positioning. The classification task is implemented on an embedded FPGA accelerator, performing high-throughput binary operations. Classification can take place at up to $\sim$6400 frames-per-second and 2W power consumption, easily enabling multi-camera and speed-gate settings. When deployed on a single entrance or gate, the idle power consumption is reduced to 1.65W, improving the battery-life of the device. We achieve an accuracy of up to 98\% for four wearing positions of the MaskedFace-Net dataset. To maintain equivalent classification accuracy for all face structures, skin-tones, hair types, and mask types, the algorithms are tested for their ability to generalize the relevant features over a diverse set of examples using the Grad-CAM approach.

\end{abstract}

%%%%%%%%%%%%%%%%%%%%%%%%%%%%%%%%%%%%%%%%%%%%%%%%%%%%%%%%%%%%%%%%%%%%%%%%%%%%%%%%
%       ___  __   __   __        __  ___    __       
%| |\ |  |  |__) /  \ |  \ |  | /  `  |  | /  \ |\ | 
%| | \|  |  |  \ \__/ |__/ \__/ \__,  |  | \__/ | \|
%%%%%%%%%%%%%%%%%%%%%%%%%%%%%%%%%%%%%%%%%%%%%%%%%%%%%%%%%%%%%%%%%%%%%%%%%%%%%%%%
\section{Introduction}
Convolutional neural networks (CNNs) have been applied to real-world problems since the early days of their conception~\cite{LeNet}. In current times, the ongoing COVID-19 pandemic presents new challenges, which can be solved with the help of state-of-the-art computer vision algorithms~\cite{covidnet, CoroNet}. One of the most simple ways of mitigating the spread of the COVID-19 disease is wearing a face-mask, which can protect the wearer from direct exposure to the virus through the mouth and nasal passages. A correctly worn mask can also protect other people, in case the wearer is already infected with the disease. This bi-directional protection makes masks highly effective in crowded and/or indoor areas. Although face-masks have become a mandatory requirement in many public areas, it is difficult to ensure the compliance of the general public. More specifically, it is difficult to assert that the masks are worn correctly as intended, \ie completely covering the nose, mouth and chin~\cite{WHO}. 

CNNs are the current state-of-the-art in face detection applications. Compared to classical computer vision algorithms, CNNs can provide better accuracy on problems with diverse features without having to manually extract said features~\cite{tradCVvsAI}. This holds true only when the training dataset has a fair distribution of samples. 
Correctly identifying a mask on a person's face is a relatively simple task for these powerful algorithms. 
However, a more precise classification of the exact positioning of the mask and identifying the exposed region of the face is more challenging. To maintain equivalent classification accuracy for all face structures, skin-tones, hair types, and mask types, the algorithms must be able to generalize the relevant features over all individuals.

The deployment scenarios for the CNN should also be taken into consideration. A face-mask detector can be set at the entrance of corporate buildings, shopping areas, airport checkpoints, and speed gates. These distributed settings require cheap, battery-powered, edge devices which are limited in memory and compute power. To maintain security and data privacy of the public, all processing must remain on the edge-device without any communication with cloud servers.

Minimizing power and resource utilization while maintaining a high classification accuracy is a design challenge which necessitates hardware-software co-design. 
In this context, we propose~\bincop{} (Binary COVID-mask Predictor), 
an efficient binary neural network (BNN) classifier for real-time classification of correct face-mask wear and positioning.
The challenges of the described application are tackled through the following contributions:
\begin{compactitem}
\item Training BNNs on synthetically generated data~\cite{maskedfacenet} to cover a wide demographic and generalize relevant task-related features. A high accuracy of $\sim$98\% is achieved for a 4-class problem of mask wear and positioning on the MaskedFace-Net dataset.
\item Deploying BNNs on a low-power, real-time embedded FPGA accelerator based on the Xilinx FINN architecture~\cite{FINN}. The accelerator can idle at a low-power of 1.65W on single entrances and gates or operate at high-performance ($\sim$6400 frames-per-second) in crowded multi-gate settings, requiring $\sim$2W of power.
\item The BNNs are analyzed through Gradient-weighted Class Activation Mapping (Grad-CAM) to improve interpretability and study the features being learned.
\end{compactitem}

%%%%%%%%%%%%%%%%%%%%%%%%%%%%%%%%%%%%%%%%%%%%%%%%%%%%%%%%%%%%%%%%%%%%%%%%%%%%%%%%
% __   ___           ___  ___  __           __   __       
%|__) |__  |     /\   |  |__  |  \    |  | /  \ |__) |__/ 
%|  \ |___ |___ /~~\  |  |___ |__/    |/\| \__/ |  \ |  \ 
%%%%%%%%%%%%%%%%%%%%%%%%%%%%%%%%%%%%%%%%%%%%%%%%%%%%%%%%%%%%%%%%%%%%%%%%%%%%%%%%
\section{Related Work}
\subsection{COVID-19 Face-Mask Wear and Positioning}
\label{sec:related_covid}
Correctly worn masks play a pivotal role in mitigating the spread of the COVID-19 disease during the ongoing pandemic~\cite{FaceMaskGer}. Members of the general public often underestimate the importance of this simple yet effective method of disease prevention and control. Researchers and data scientists in the field of computer vision have collected data to train and deploy algorithms which help in automatically regulating masks in public spaces and indoor locations~\cite{MAFA, wang2020masked}. Although large-scale natural face datasets exist, the number of real-world masked images is limited~\cite{MAFA}. Wang et al.~\cite{wang2020masked} extended their masked-face dataset with a Simulated Masked Face Recognition Dataset (SMFRD), which is synthetically generated by applying virtual masks to existing natural face datasets. Cabani et al.~\cite{maskedfacenet} improved the generation of synthetically masked-faces by applying a deformable mask-model onto natural face images with the help of automatically detected facial key-points. The key-points of the deformable mask-model can be matched to the key-points of the face, allowing the application of the mask in a variety of ways. This allows the dataset generation process to further generate examples of incorrectly worn masks, such as chin exposed, nose exposed or nose and mouth exposed.

\subsection{Binary Neural Networks}
The memory footprint of neural networks and the complexity of their arithmetic operations on inference hardware can be reduced through parameter quantization. In the most extreme case, binarizing neural networks constrains their weights and activations to $\{-1,1\}$, such that their memory footprint is theoretically reduced by $\times$32 compared to a float-32 CNN~\cite{BNN}. Additionally, simple  $\mathtt{XNOR}$ and $\mathtt{popcount}$ operations can be used to implement multiply-accumulate (MAC) operations on inference hardware~\cite{XNOR}. Specialized training schemes have been proposed to mitigate the loss in information capacity introduced by the low-bitwidth representation of BNNs~\cite{BNN, BinaryConnect, ABCNET, XNOR}.  
In some cases, the low information capacity due to binarization can have a regularization effect which improves feature generalization~\cite{BinaryConnect}. This is helpful in improving the classification performance on real-world data, particularly when training on synthetically generated data~\cite{bdad_fri_20b}.
In~\cite{BinaryConnect}, Courbariaux~\etal introduced a scheme to train neural networks with binary weights during forward propagation while maintaining latent full-precision values during back propagation. This ensures proper gradient flow and fine adjustments through the gradients. This approach is later extended by the binarization of activations~\cite{BNN}. Rastegari~\etal~\cite{XNOR} proposed XNOR-Net, where both weights and activations are binarized such that the convolutions of input feature maps and weights can be approximated by a combination of $\mathtt{XNOR}$ operations and $\mathtt{popcounts}$, followed by a multiplication with scaling factors. The introduction of scaling factors improves the information capacity of the network at the cost of more trainable parameters for each layer. This adds to the computational complexity of XNOR-Net at deployment time. 
For the task of face-mask detection with a single subject in the frame (e.g. gates and entrance points), more efficient forms of BNNs~\cite{BNN} can be applied.

\subsection{BNN Hardware Accelerators}
\label{sec:related_hw}
Several accelerators have been designed to exploit the benefits of BNNs \cite{7878541, BRein, FINN, towardshwBNN}. The Xilinx FINN \cite{FINN} framework was developed to accelerate BNNs efficiently on FPGA platforms. The framework compiles high level synthesis (HLS) code from a BNN description to create a hardware design for the network. The generated streaming architecture consists of a pipeline of individual hardware components instantiated for each layer of the BNN. 
In this work, we deploy~\bincop{} on FINN-based hardware architectures to achieve an efficient acceleration of the masked-face inference on embedded FPGAs. We parameterize and synthesize accelerators with different hardware requirements, geared towards individual COVID-19 mask recognition (low-power) or multi-camera (multi-gate) classification (high-performance).

%%%%%%%%%%%%%%%%%%%%%%%%%%%%%%%%%%%%%%%%%%%%%%%%%%%%%%%%%%%%%%%%%%%%%%%%%%%%%%%%
%      ___ ___       __   __  
%|\/| |__   |  |__| /  \ |  \ 
%|  | |___  |  |  | \__/ |__/ 
%%%%%%%%%%%%%%%%%%%%%%%%%%%%%%%%%%%%%%%%%%%%%%%%%%%%%%%%%%%%%%%%%%%%%%%%%%%%%%%%
\section{Method}
\begin{figure*}[t]
\centering
\resizebox{0.95\linewidth}{!}{
	\includegraphics{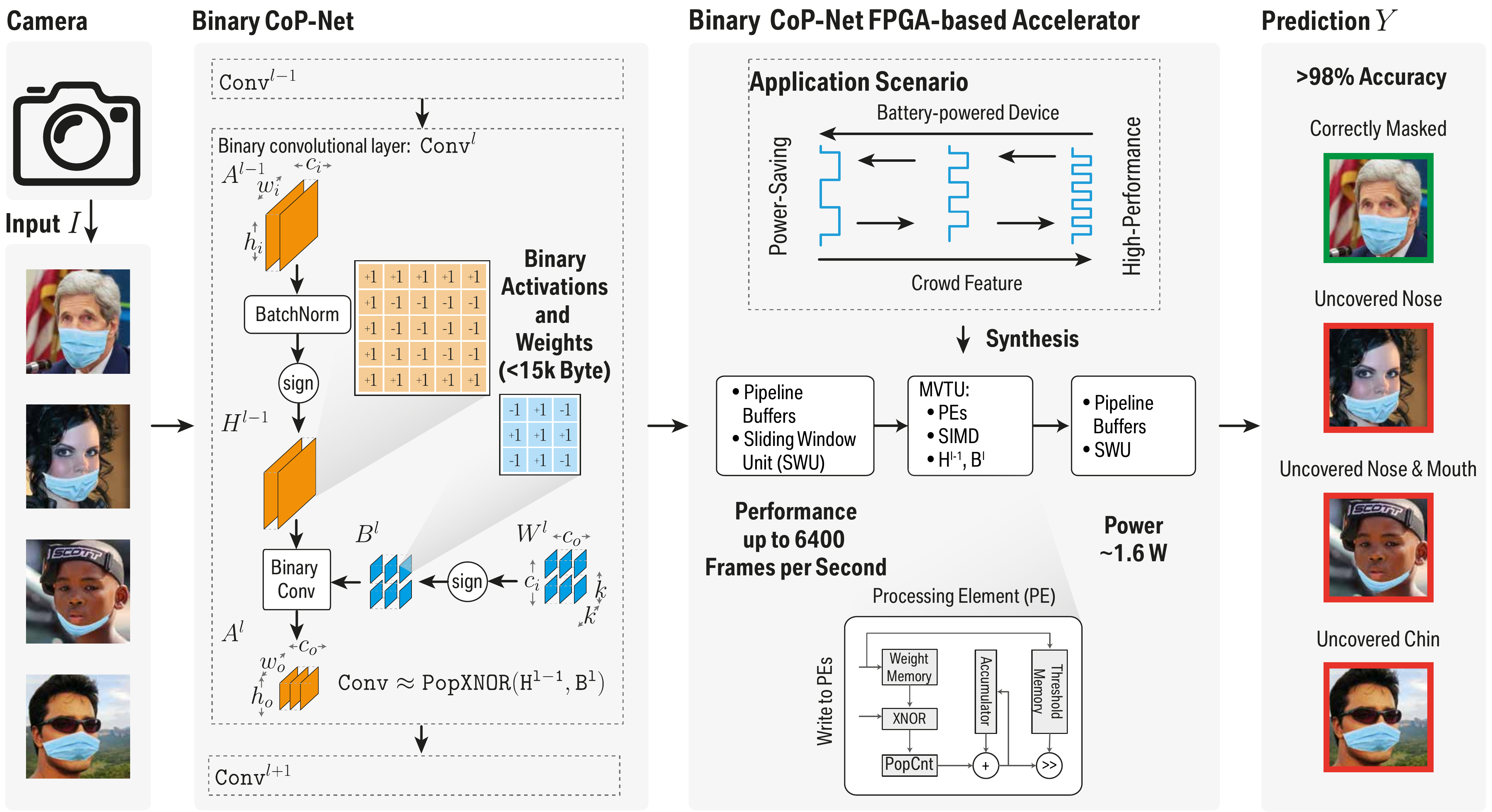}}
	\caption{Schematic representation of the \bincop{} accelerator. A camera captures images to be classified by the neural network. The BNN accelerator is tailored for the application scenario (single or multi-gate prediction). Binary tensors are processed in the PEs of the FPGA-based accelerator using efficient XNOR operations.}
	\label{fig:bincop}
	\vspace{-3ex}
\end{figure*}
\subsection{Training and Inference of Binary Neural Networks}
\label{sec:method_bnn}
The BNN method proposed by Courbariaux~\etal~\cite{BNN} serves as our foundation to efficiently approximate weights and activations to single-bit precision at inference time, such that the neural network's arithmetic operations can be executed as simple logic operations. Smooth model training and convergence is ensured by relying on full-precision latent weights $W$ during training time~\cite{STE}. In detail, the activation tensor $A^{l-1}\in \mathbb{R}^{X_i\times Y_i\times C_i}$, with its dimensions of $X_i$ width, $Y_i$ height, and $C_i$ channels, serves as the input to the convolutional layer  $l\in[1,...,L]$. Here, $A^{0}$ and $A^{L}$ represent the input image and the network's prediction, respectively. The trainable parameters of the 2D-convolutional layers are composed of the latent weight matrix $W\in \mathbb{R}^{K\times K\times C_i\times C_o}$ required for training, with kernel dimension $K$, input channels $C_i$, and output channels $C_o$. As previously stated, the latent weights are mapped to $\{-1, +1\}$ during the forward pass for loss calculation or deployment, resulting in the binarized $b \subset B\in \mathbb{B}^{K\times K\times C_i \times C_o}$. In the hardware implementation, $-1$ is expressed as a binary $0$ to perform multiplications as $\mathtt{XNOR}$ logic operations.
The $\mathtt{sign}()$ function in Eq.~\ref{eq:sign} is used to binarize the input feature maps and weights.

\begin{equation}
\label{eq:sign}
    b = \mathtt{sign}(w)  = \bigg\{\begin{array}{ll}
             1 & \text{if}~w \geq 0,\\
            -1 & \text{otherwise} \\ 
\end{array}.
\end{equation}
The derivative of the $\mathtt{sign}()$ function is almost always zero, resulting in insufficient gradient flow during training and back-propagation. This necessitates gradient flow approximation using a straight-through estimator (STE)~\cite{STE}. 

Particularly for BNNs, it is of crucial importance to adjust the input elements $a^{l-1} \subset A^{l-1}$, before the approximation into the binary representation $h^{l-1} \subset H^{l-1}\in\mathbb{B}^{X_i\times Y_i\times C_i}$ by means of batch normalization to zero mean and unit variance. An advantage of BNNs is that the result of the batch-norm operation is followed by $\mathtt{sign}()$ (see Fig.~\ref{fig:bincop}). Since the result after applying both functions is simply $\{-1,1\}$, the precise calculation of the batch-norm is wasteful on embedded hardware. Based on the batch-norm statistics collected at training time, a \textit{threshold} point $\tau$ is defined, wherein an activation value $a^{l-1}\geq\tau$ results in 1, otherwise \mbox{-1}~\cite{FINN}. This allows the implementation of the typically costly batch-norm operation as a simple magnitude comparison operation on hardware.

Next, the binary convolution follows as:
\begin{equation}
    H^{l-1}=\mathtt{sign}(\mathtt{BatchNorm}(A^{l-1})); B^l=\mathtt{sign}(W^l)
\end{equation}
\begin{equation}
    A^l=\mathtt{BinConv}(H^{l-1},B^l)=\mathtt{PopCnt}(\mathtt{XNOR}(H^{l-1},B^l)),
\end{equation}
which results in the output feature map $A^l \in \mathbb{R}^{X_o\times Y_o\times C_o}$.

\subsection{Hardware Architecture}
The trained BNNs are conditioned for deployment on the Xilinx FINN framework~\cite{FINN}. The pipelined architecture offers several advantages on embedded devices, most importantly, the reduction in on-chip to off-chip memory transfers of the BNN parameters $B^{l}$ and intermediate activations $A^{l}$ and $H^{l}$. This is mainly feasible due to the binary format, which results in highly compact neural networks that can fit on the on-chip memory units of embedded devices. The number of processing elements (PEs), single-instruction-multiple-data (SIMD)-lanes, and other parameters can be optimized by the designer to suit the acceleration of the trained BNN. The final design is synthesized and implemented on an embedded FPGA.

For each convolutional or fully-connected layer in the BNN, a matrix-vector-threshold unit (MVTU) is instantiated, which executes the $\mathtt{XNOR}$, $\mathtt{popcount}$ and threshold operations mentioned in Sec.~\ref{sec:method_bnn}. Each MVTU in the pipeline can be dimensioned for the number of PEs and SIMD lanes, which have a significant impact on hardware resource utilization, latency and the effective throughput of the pipeline. Based on the compute complexity of each layer, the available hardware resources need to be distributed over the corresponding MVTUs, such that all parts of the pipeline have a matched-throughput. A single under-dimensioned MVTU could throttle the entire pipeline, resulting in sub-optimal throughput. A single MVTU of the pipeline is shown in Fig.~\ref{fig:bincop}, and a corresponding PE is detailed.

For convolutional layers, an additional sliding-window unit (SWU) reshapes the binarized activation maps to create a single, wide input feature map memory, which can efficiently be accessed by the corresponding MVTU. Max-pool layers are implemented as boolean $\mathtt{OR}$ operations, since a single binary ``1'' value suffices to make the entire pool window output equal to 1.

\subsection{BNN Interpretability with Grad-CAM}
The output of the convolutional layers in a CNN contains localized information of the input image, without any prior bias on the location of objects and features during training. This information can be captured using Class Activation Mapping (CAM)~\cite{cam} and Gradient-weighted Class Activation Mapping (Grad-CAM)~\cite{Selvaraju_2017_ICCV} techniques. To apply CAM, the model must end with a global average pooling layer followed by a fully-connected layer, providing the logits of a particular input. The BNN models investigated in this work operate on a small input resolution of 32$\times$32, and achieve a high reduction of spatial information without incorporating a global average pooling layer. For this reason, the Grad-CAM approach is better-suited to obtain visual interpretations of \bincop{}'s attention and determine the important regions for its predictions of different inputs and classes. 

To obtain the class-discriminative localization map, we consider the activations and gradients for the output of the \texttt{Conv2\_2} layer (see Tab.~\ref{tab:bnn_arch}), which has spatial dimensions of 5$\times$5. We use average pooling for the corresponding gradients and reduce the channels by performing Einstein summation as specified in~\cite{Selvaraju_2017_ICCV}. With this approach the base networks do not need any modifications or retraining. Due to the synthetically generated dataset used for training, we expect \bincop{} models to generalize well against domain shifts. 

%%%%%%%%%%%%%%%%%%%%%%%%%%%%%%%%%%%%%%%%%%%%%%%%%%%%%%%%%%%%%%%%%%%%%%%%%%%%%%%%
% ___      __   ___  __           ___      ___  __  
%|__  \_/ |__) |__  |__) |  |\/| |__  |\ |  |  /__` 
%|___ / \ |    |___ |  \ |  |  | |___ | \|  |  .__/
%%%%%%%%%%%%%%%%%%%%%%%%%%%%%%%%%%%%%%%%%%%%%%%%%%%%%%%%%%%%%%%%%%%%%%%%%%%%%%%%
\section{Results and Design Space Exploration}
\subsection{Experimental Setup}
\begin{table}[b]
    \begin{center}
	\caption{Network architectures and hardware dimensioning.}
	\label{tab:bnn_arch}
	    \resizebox{\linewidth}{!}{
        \begin{tabular}{l|c|c|c}
        \toprule
        \textbf{Network} & \textbf{CNV}& \textbf{$n$-CNV} & \textbf{$\mu$-CNV} \\
        \midrule
        \midrule
        \makecell[tl]{\textbf{Arch.} \\  $L~|~[C_{i},C_{o}]$ \\ $K = 3~\forall$ Conv}  & \makecell[tl]{\texttt{Conv1\_1} $|$ [3, 64] \\ \texttt{Conv1\_2} $|$ [64, 64] \\ \texttt{Conv2\_1} $|$ [64, 128] \\
        \texttt{Conv2\_2} $|$ [128, 128] \\ \texttt{Conv3\_1} $|$ [128, 256] \\ \texttt{Conv3\_2} $|$ [256, 256] \\
        \texttt{FC1} $|$ [512] \\ \texttt{FC2} $|$ [512] \\ \texttt{FC3} $|$ [4]} & 
        
        \makecell[tl]{\texttt{Conv1\_1} $|$ [3, 16] \\ \texttt{Conv1\_2} $|$ [16, 16] \\ \texttt{Conv2\_1} $|$ [16, 32] \\
        \texttt{Conv2\_2} $|$ [32, 32] \\ \texttt{Conv3\_1} $|$ [32, 64] \\ \texttt{Conv3\_2} $|$ [64, 64] \\
        \texttt{FC1} $|$ [128] \\ \texttt{FC2} $|$ [128] \\ \texttt{FC3} $|$ [4]} & 
        
        \makecell[tl]{\texttt{Conv1\_1} $|$ [3,16] \\ \texttt{Conv1\_2} $|$ [16, 16] \\ \texttt{Conv2\_1} $|$ [16, 32] \\
        \texttt{Conv2\_2} $|$ [32, 32] \\ \texttt{Conv3\_1} $|$ [32, 64] \\ \texttt{FC1} $|$ [128] \\ \texttt{FC2} $|$ [4]}  \\ 
        \midrule
        \textbf{PE Count} & 16, 32, 16, 16, 4, 1, 1, 1, 4 & 16, 16, 16, 16, 4, 1, 1, 1, 1 & 4, 4, 4, 4, 1, 1, 1 \\
        \textbf{SIMD lanes} & 3, 32, 32, 32, 32, 32, 4, 8, 1 & 3, 16, 16, 32, 32, 32, 4, 8, 1 & 3, 16, 16, 32, 32, 16, 1 \\
 		\bottomrule
    \end{tabular}}
    \vspace{-3ex}
    \setlength{\belowcaptionskip}{-25pt}
   \end{center}
\end{table}

\bincop{} is able to detect the presence of a mask, as well as its position and correctness. This level of classification detail is possible through the more detailed split of the MaskedFace-Net dataset~\cite{maskedfacenet} from 2 classes, namely Correctly Masked Face Dataset (CMFD) and Incorrectly Masked Face Dataset (IMFD), to 4 classes of CMFD, IMFD Nose, IMFD Chin, and IMFD Nose and Mouth. The dataset suffers from high imbalance in the number of samples per class. From the total 133,783 samples, roughly 5\% of the samples are IMFD Chin, and another 5\% samples are IMFD Nose and Mouth. CMFD samples make up 51\% of the total dataset while IMFD Nose makes up 39\%. The dataset in its raw distribution would heavily bias the training towards the two dominant classes. To counter this, we randomly sample  the larger classes CMFD and IMFD Nose to collect a comparable number of examples to the two remaining classes, IMFD Chin and IMFD Nose and Mouth. The evenly balanced dataset is then randomly augmented with a varying combination of contrast, brightness, gaussian noise, flip and rotate operations. The final size of the balanced dataset is 110K train and validation examples and 28K test samples. The images are resized to 32$\times$32 pixels, similar to the CIFAR-10~\cite{cifar10} dataset. The BNNs are trained up to 300 epochs, unless learning saturates earlier. The full-precision (FP32) variant used for the Grad-CAM comparison is trained for 175 epochs due to early learning saturation (98.6\% final test accuracy). 
We trained the BNN architectures shown in Tab.~\ref{tab:bnn_arch} according to the method described in Sec.\ref{sec:method_bnn}. Each convolutional (\texttt{Conv}) and fully-connected (\texttt{FC}) layer is followed by batch-norm and activation layers except for the final layer. \texttt{Conv} groups \texttt{1} and \texttt{2} are followed by a max-pool layer. 
The target System-on-Chip (SoC) platform for the experiments is the Xilinx XC7Z020 (Z7020) chip on the PYNQ-Z1 board. The $\mu$-CNV design can also be synthesized for the more constrained XC7Z010 (Z7010) chip, when $\mathtt{XNOR}$ operations are offloaded to the DSP blocks as described in~\cite{orthruspe}. Power and throughput measurements are taken directly on a running system. The power is measured at the power supply of the board (includes both the processing system and programmable logic). The throughput reported is the classification rate when the accelerator's pipeline is full, while latency is measured end-to-end for a single image entering and exiting the pipeline. It should be noted that due to the pipelined architecture, throughput is not simply the reciprocal of latency. When the pipeline is full, each MVTU (layer) ideally operates on a different image of the input batch, \ie concurrently processing $L$ images at different stages of the accelerator.
\subsection{Design Space Exploration}
\iffalse
\begin{itemize}
    \item $\mu$-CNV and $m$-CNV targeted at Z7010. \textit{Cannot} be synthesized without moving XNORs to DSPs.
    \item All configurations have reaction time below 90ms in low-power mode. Max latency can be chosen according to delay of other parts of the system
    \item Throughput for $m$-CNV does not scale, due to bottleneck in first layers, can be solved if more PEs were synthesizable on Z7010
    \item $\mu$-CNV suffers same problem, but less critical as it has some parallelism in initial layers. Could be improved if all layers were equaly sharing the total delay (not the case here)
    \item Confusion matrix, CIFAR accuracy if possible (and enought space left)
\end{itemize}
\fi

\begin{table}[t]
    \begin{center}
	\caption{Hardware results of design space exploration.}
	\label{tab:dse_results}
	\large
	    \resizebox{\linewidth}{!}{
        \begin{tabular}{l|ccc|c|c|c|c|c}
        \toprule
        \multirow{2}{*}{\textbf{Prototype}} & \multirow{2}{*}{\textbf{LUT}} & \multirow{2}{*}{\textbf{BRAM}} & \multirow{2}{*}{\textbf{DSP}} & \multicolumn{2}{c|}{\textbf{Power} [W]} & \textbf{Thr.put} & \textbf{Latency} & \textbf{Acc.} \\
        & & & & Idle & Inf. & [FPS] & [ms] & [\%]\\
		\midrule
		\midrule
		CNV & 26060 & 124 & 24 &  & 2.212 & 3049 & 1.58 & \textbf{98.10} \\
		\cmidrule{1-4} \cmidrule{6-9}
		$n$-CNV & 20425 & \textbf{10.5} &14 & 1.65* & 2.122 & \textbf{6460} & 0.31 & 93.94 \\ 
		\cmidrule{1-4} \cmidrule{6-9}
		$\mu$-CNV & \textbf{11738} & 14 & 27 & & 2.028 & 1646 & 0.81 & 93.78 \\
 		\bottomrule
 		\multicolumn{9}{r}{*Required by the board and ARM-Cortex A9 processor. Accelerator is idle.}
    \end{tabular}}
    \setlength{\belowcaptionskip}{-25pt}
   \end{center}
   \vspace{-4.25ex}
\end{table}

We evaluate three \bincop{}~prototypes, namely CNV, $n$-CNV and $\mu$-CNV. The CNV network is based on the architecture in~\cite{FINN} inspired by VGG-16~\cite{VGG} and BinaryNet~\cite{BNN}. $n$-CNV is a downsized version for a smaller memory footprint, and $\mu$-CNV has fewer layers to reduce the size of the synthesized design. All designs are synthesized with a target clock frequency of 100MHz.

\begin{figure}[t]
\begin{tikzpicture}
		\begin{axis}[
		name=OPs,
		height=0.6\columnwidth,
		width=\columnwidth,
		ybar,
		bar width= 5pt,
		bar shift=-2.5pt,
		grid,
		axis y line*=left,
		xtick pos=left,
        ytick pos=left,
  		legend style={at={(0.6,0.85)},
  		anchor=west,legend columns=-1, draw=none, nodes={scale=0.7, transform shape}, column sep=1pt},
  		ylabel={Binary OPs ($10^6$)},
		symbolic x coords={\texttt{Conv1\_1}, \texttt{Conv1\_2}, \texttt{Conv2\_1}, \texttt{Conv2\_2}, \texttt{Conv3\_1}, \texttt{Conv3\_2}, \texttt{FC1}, \texttt{FC2}, \texttt{FC3}},
		xtick=data,
		xticklabel style = {rotate=45, font=\scriptsize, yshift=1ex},
		yticklabel style = {rotate=90, font=\scriptsize},
		ylabel style = {font=\scriptsize, yshift=-4ex},
		ymin=0,
		ymax=4,
        enlarge x limits={0.1},
		%nodes near coords,
		%nodes near coords align={vertical},
		%cycle list name=colorlist1,
		%postaction={pattern=horizontal lines, very thick,pattern color=white},
		]
		%nCNV
		\addplot[fill=blue!50!white] coordinates {(\texttt{Conv1\_1}, 0.777600) (\texttt{Conv1\_2}, 3.612672) (\texttt{Conv2\_1}, 1.327104) (\texttt{Conv2\_2}, 1.843200) (\texttt{Conv3\_1}, 0.331776) (\texttt{Conv3\_2}, 0.073728) (\texttt{FC1}, 0.016384) (\texttt{FC2}, 0.032768) (\texttt{FC3}, 0.016384) };

 		\end{axis}
 		
 		\begin{axis}[
		name=Latency,
		height=0.6\columnwidth,
		width=\columnwidth,
		ybar,
		bar width= 5pt,
		bar shift=2.5pt,
		ymajorgrids=true,
  		legend style={at={(0.65,0.92)},
  		anchor=west,legend columns=-1, draw=none, nodes={scale=0.7, transform shape}, column sep=1pt},
  		ylabel={Est. Latency ($10^3$Cycles)},
  		tick pos = right,
  		%yticklabel pos=right,
  		ylabel near ticks,
  		axis y line*=right,
  		axis x line=none,
		symbolic x coords={\texttt{Conv1\_1}, \texttt{Conv1\_2}, \texttt{Conv2\_1}, \texttt{Conv2\_2}, \texttt{Conv3\_1}, \texttt{Conv3\_2}, \texttt{FC1}, \texttt{FC2}, \texttt{FC3}},
		xtick=data,
		xticklabels=\empty,
		xticklabel style = {rotate=45, font=\scriptsize, yshift=1ex},
		yticklabel style = {rotate=90, font=\scriptsize},
		ylabel style = {font=\scriptsize, yshift=0.5ex},
		ymin=0,
        ymax=10,
        ytick={0,2.5,5,7.5,10},
        enlarge x limits={0.1},
		%nodes near coords,
		%nodes near coords align={vertical},
		%cycle list name=colorlist1,
		%postaction={pattern=horizontal lines, very thick,pattern color=white},
		]
		%nCNV
		\addplot[forget plot, fill=red!50!white] coordinates {(\texttt{Conv1\_1}, 8.100) (\texttt{Conv1\_2}, 7.056) (\texttt{Conv2\_1}, 2.592) (\texttt{Conv2\_2}, 1.800) (\texttt{Conv3\_1}, 1.296) (\texttt{Conv3\_2}, 1.152) (\texttt{FC1}, 2.048) (\texttt{FC2}, 2.048) (\texttt{FC3}, 8.192) }; 
		
		\addlegendimage{fill=blue!50!white}
		\addlegendentry{OPs}
		\addlegendimage{fill=red!50!white}
		\addlegendentry{Latency}
		
		%\draw[red, line width=1pt] (20, 80) circle [x radius=2.5, y radius=0.5];
		\node[align=center] at (rel axis cs: 0.37, 0.85) {\footnotesize Highest OPs \\ \footnotesize $\rightarrow$ Throughput Setter};
		
		\node[align=center] at (rel axis cs: 0.70
		, 0.4) {\footnotesize Low OPs \\ \footnotesize High Memory};
		
		\draw[->] (rel axis cs: 0.59, 0.05) -- (rel axis cs: 0.59, 0.3);
		
		\draw[->] (rel axis cs: 0.69, 0.05) -- (rel axis cs: 0.69, 0.3);
		
		\draw[->] (rel axis cs: 0.795, 0.05) -- (rel axis cs: 0.795, 0.3);
		
		\draw[->] (rel axis cs: 0.9, 0.25) -- (rel axis cs: 0.85, 0.35);
		
		\draw[<->] (rel axis cs: 0.9, 0.72) -- (rel axis cs: 0.25, 0.72);
		
		\node[align=center] at (rel axis cs: 0.75
		, 0.725) {\footnotesize Rate matched \\ \footnotesize Less PEs};
		
 		\end{axis}
 		
		\end{tikzpicture}´
		\caption{Binary operations and layer-wise latency estimates based on PE/SIMD choices for \bincop{}-$n$-CNV.}
	\label{fig:lat_ops_ncnv}
	\vspace{-4.25ex}
\end{figure}
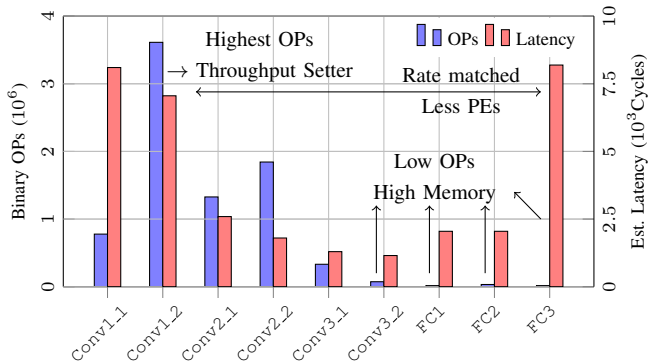

Referring back to Tab.~\ref{tab:bnn_arch}, the PE counts and SIMD-lanes for each layer (MVTU) are shown in sequence. For \bincop{}-$n$-CNV, the most complex layer is \texttt{Conv1\_2} with 3.6M $\mathtt{XNOR}$ and $\mathtt{popcount}$ operations. In Fig.~\ref{fig:lat_ops_ncnv}, we mark this layer as the throughput setter, due to its heavy influence on the final throughput of the accelerator. Allocating more PEs for this layer's MVTU increases the overall throughput of the pipeline, so long as no other layer becomes the bottleneck. We allocate enough resources for \texttt{Conv1\_1} to roughly match \texttt{Conv1\_2}'s latency. The FINN architecture employs a weight-stationary dataflow, since each PE has its own pre-loaded weight memory. When the total number of parameters of a given layer increases, it becomes important to map these parameters to BRAM (Block RAM) units instead of logic. The deeper layers have several orders of magnitude fewer OPs, but more parameters. For these layers, increasing the number of PEs fragments the total weight memory, leading to worse BRAM utilization and no benefit in terms of throughput. Here, choosing fewer PEs, with larger unified weight memories, leads to improved memory allocation, while maintaining rate-matching with the shallow layers (see Fig.~\ref{fig:lat_ops_ncnv}), leaving the throughput gains from the initial PEs unhindered. The CNV architecture in ~\cite{FINN} follows the same reasoning for PE and SIMD allocation. For $\mu$-CNV, we choose fewer PEs for the throughput-setters, as this prototype is meant to fit on embedded FPGAs with less emphasis on high frame rates.

%pipeline is the MVTU of the first layer which has 16 PEs with 3 SIMD lanes each. This is due to the higher total number of operations required in the shallow parts of network ($\sim$389K binary MAC ops for Conv\_1). This results in an estimated latency of 32400 cycles for this layer.

In Tab.~\ref{tab:dse_results}, the hardware utilization for the \bincop{} prototypes is provided. With $\mu$-CNV, a significant reduction in LUTs is achieved, which makes the design synthesizable on the heavily constrained Z7010 SoC. The trade-off is a slight increase in the memory footprint of the BNN, as the shallower network has a larger spatial dimension before the fully-connected layers, increasing the total number of parameters after the last convolutional layer. The choice of PE count and SIMD lanes for the $n$-CNV prototype allow it to reach a maximum throughput of $\sim$6400 classifications per second when its pipeline is full. This high-performance can be used to classify images from multiple cameras in multi-gate settings. %Alternatively, a face detector can pass many cropped images from a crowded scene to \bincop{} for statistics collection. to  
The inference power values reported in Tab.~\ref{tab:dse_results} show a total power requirement of around 2W for all prototypes. For single entrance/gate classifications, all prototypes have an idle power of around 1.65W. In this setting, a classification needs to be triggered only when a subject is attempting to pass through the entrance where \bincop{} is deployed. The idle power is required mostly by the processor (ARM-Cortex A9) on the SoC and the board (PYNQ-Z1). This can be reduced further by choosing a smaller processor to pair with the proposed hardware accelerator. Although the PYNQ-Z1 board has no PMBus to isolate the power measurements of the FPGA from the rest of the components, we can infer that the hardware accelerator requires roughly 0.4W for the inference task from the two measured power values in Tab.~\ref{tab:dse_results}. The current design is still dependent on the processor for pre- and post-processing, therefore we report the joint power for fairness.

\begin{figure}[t]
    \centering
    %\vspace{2ex}
    \resizebox{0.8\linewidth}{!}{
    %The matrix in numbers
%Horizontal target class
%Vertical output class
\def\myConfMat{{
{7125,41,1,90},  %row 1
{26,7042,94,26},  %row 2
{4,79,5651,9},  %row 3
{107,41,7,7363},  %row 4
}}

\def\classNames{{"Correct","Nose","N+M","Chin"}} %class names. Adapt at will

\def\numClasses{4} %number of classes. Could be automatic, but you can change it for tests.

\def\myScale{1.5} % 1.5 is a good scale. Values under 1 may need smaller fonts!
\begin{tikzpicture}[
    scale = \myScale,
    %font={\scriptsize}, %for smaller scales, even \tiny may be useful
    ]

\tikzset{vertical label/.style={rotate=90,anchor=east}}   % usable styles for below
\tikzset{diagonal label/.style={rotate=45,anchor=north east}}

\foreach \y in {1,...,\numClasses} %loop vertical starting on top
{
    % Add class name on the left
    \node [anchor=east] at (0.4,-\y) {\pgfmathparse{\classNames[\y-1]}\pgfmathresult}; 
    
    \foreach \x in {1,...,\numClasses}  %loop horizontal starting on left
    {
%---- Start of automatic calculation of totSamples for the column ------------   
    \def\totSamples{0}
    \foreach \ll in {1,...,\numClasses}
    {
        \pgfmathparse{\myConfMat[\ll-1][\x-1]}   %fetch next element
        \xdef\totSamples{\totSamples+\pgfmathresult} %accumulate it with previous sum
        %must use \xdef fro global effect otherwise lost in foreach loop!
    }
    \pgfmathparse{\totSamples} \xdef\totSamples{\pgfmathresult}  % put the final sum in variable
%---- End of automatic calculation of totSamples ----------------
    
    \begin{scope}[shift={(\x,-\y)}]
        \def\mVal{\myConfMat[\y-1][\x-1]} % The value at index y,x (-1 because of zero indexing)
        \pgfmathtruncatemacro{\r}{\mVal}   %
        \pgfmathtruncatemacro{\p}{round(\r/\totSamples*100)}
        \coordinate (C) at (0,0);
        \ifthenelse{\p<50}{\def\txtcol{black}}{\def\txtcol{white}} %decide text color for contrast
        \node[
            draw,                 %draw lines
            text=\txtcol,         %text color (automatic for better contrast)
            align=center,         %align text inside cells (also for wrapping)
            fill=black!\p,        %intensity of fill (can change base color)
            minimum size=\myScale*10mm,    %cell size to fit the scale and integer dimensions (in cm)
            inner sep=0,          %remove all inner gaps to save space in small scales
            ] (C) {\r\\\p\%};     %text to put in cell (adapt at will)
        %Now if last vertical class add its label at the bottom
        \ifthenelse{\y=\numClasses}{
        \node [] at ($(C)-(0,0.75)$) % can use vertical or diagonal label as option
        {\pgfmathparse{\classNames[\x-1]}\pgfmathresult};}{}
    \end{scope}
    }
}
%Now add x and y labels on suitable coordinates
\coordinate (yaxis) at (-0.5,0.15-\numClasses/2);  %must adapt if class labels are wider!
\coordinate (xaxis) at (0.5+\numClasses/2, -\numClasses-1.25); %id. for non horizontal labels!
\node [vertical label] at (yaxis) {True Class};
\node []               at (xaxis) {Predicted Class};
\end{tikzpicture}
    }\hspace{6ex}
    \caption{Confusion matrix of \bincop-CNV on the test set.}
    \vspace{-4ex}
    \label{fig:confusion}
\end{figure}
\subsection{Grad-CAM and Confusion Matrix Analysis}
The confusion matrix in Fig.~\ref{fig:confusion} shows the generalization of \bincop{-}CNV on all classes after balancing the dataset. As expected, it is extremely rare to mistake nose+mouth exposed with a correctly worn mask. Less critically, nose and nose+mouth have a slight misclassification overlap, still at only 2\% of the total samples given for each class. Finally, the chin exposed and the correct class have some sample misclassifications ($\leq$1\%), which could be attributed to the chin area being small in some images and hard to detect at low-resolution.

\begin{figure*}[t]
\scriptsize
       \subfloat[Correctly masked Grad-CAM]{
       \begin{tabular}{c|cccc}
       \toprule
\multirow{2}{*}{\textbf{Label}} & \multirow{2}{*}{\textbf{Raw}} & \textbf{BCoP} & \textbf{BCoP} & \multirow{2}{*}{\textbf{FP32}} \\
& & CNV & $n$-CNV & \\
\midrule
\midrule    
    \raisebox{4ex}{\makecell{Correctly \\ Masked}} & \includegraphics[width=0.08\textwidth, height=0.08\textwidth]{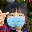}
      & \includegraphics[width=0.08\textwidth, height=0.08\textwidth]{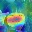}
      &
    \includegraphics[width=0.08\textwidth, height=0.08\textwidth]{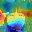}
      &
    \includegraphics[width=0.08\textwidth, height=0.08\textwidth]{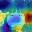}
      \\
    
    \raisebox{4ex}{\makecell{Correctly \\ Masked}} & \includegraphics[width=0.08\textwidth, height=0.08\textwidth]{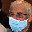}
      & \includegraphics[width=0.08\textwidth, height=0.08\textwidth]{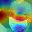}
      &
    \includegraphics[width=0.08\textwidth, height=0.08\textwidth]{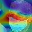}
      &
    \includegraphics[width=0.08\textwidth, height=0.08\textwidth]{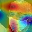}
      \\
    
    \raisebox{4ex}{\makecell{Correctly \\ Masked}} & \includegraphics[width=0.08\textwidth, height=0.08\textwidth]{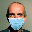}
      & \includegraphics[width=0.08\textwidth, height=0.08\textwidth]{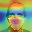}
      &
    \includegraphics[width=0.08\textwidth, height=0.08\textwidth]{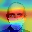}
      &
    \includegraphics[width=0.08\textwidth, height=0.08\textwidth]{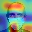}
      \\
    \bottomrule
    \end{tabular}
      }\hfill
        \subfloat[Nose exposed Grad-CAM]{
\begin{tabular}{c|cccc}
\toprule
\multirow{2}{*}{\textbf{Label}} & \multirow{2}{*}{\textbf{Raw}} & \textbf{BCoP} & \textbf{BCoP} & \multirow{2}{*}{\textbf{FP32}} \\
& & CNV & $n$-CNV & \\
\midrule
\midrule
    \raisebox{4ex}{\makecell{Nose \\ Exposed}} & \includegraphics[width=0.08\textwidth, height=0.08\textwidth]{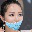}
      & \includegraphics[width=0.08\textwidth, height=0.08\textwidth]{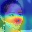}
      &
    \includegraphics[width=0.08\textwidth, height=0.08\textwidth]{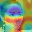}
      &
    \includegraphics[width=0.08\textwidth, height=0.08\textwidth]{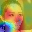}
      \\
    
    \raisebox{4ex}{\makecell{Nose \\ Exposed}} & \includegraphics[width=0.08\textwidth, height=0.08\textwidth]{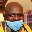}
      & \includegraphics[width=0.08\textwidth, height=0.08\textwidth]{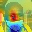}
      &
    \includegraphics[width=0.08\textwidth, height=0.08\textwidth]{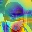}
      &
    \includegraphics[width=0.08\textwidth, height=0.08\textwidth]{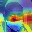}
      \\
    
    \raisebox{4ex}{\makecell{Nose \\ Exposed}} & \includegraphics[width=0.08\textwidth, height=0.08\textwidth]{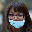}
      & \includegraphics[width=0.08\textwidth, height=0.08\textwidth]{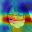}
      &
    \includegraphics[width=0.08\textwidth, height=0.08\textwidth]{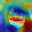}
      &
    \includegraphics[width=0.08\textwidth, height=0.08\textwidth]{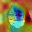}
      \\
      \bottomrule
\end{tabular}
       }
       \\
        \subfloat[Mouth + nose exposed Grad-CAM]{\begin{tabular}{c|cccc}
        \toprule
\multirow{2}{*}{\textbf{Label}} & \multirow{2}{*}{\textbf{Raw}} & \textbf{BCoP} & \textbf{BCoP} & \multirow{2}{*}{\textbf{FP32}} \\
& & CNV & $n$-CNV & \\
\midrule
\midrule
    \raisebox{4ex}{\makecell{Nose \\ Mouth \\ Exposed}} & \includegraphics[width=0.08\textwidth, height=0.08\textwidth]{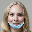}
      & \includegraphics[width=0.08\textwidth, height=0.08\textwidth]{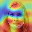}
      &
    \includegraphics[width=0.08\textwidth, height=0.08\textwidth]{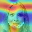}
      &
    \includegraphics[width=0.08\textwidth, height=0.08\textwidth]{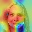}
      \\
    
    \raisebox{4ex}{\makecell{Nose \\ Mouth \\ Exposed}} & \includegraphics[width=0.08\textwidth, height=0.08\textwidth]{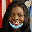}
      & \includegraphics[width=0.08\textwidth, height=0.08\textwidth]{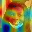}
      &
    \includegraphics[width=0.08\textwidth, height=0.08\textwidth]{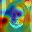}
      &
    \includegraphics[width=0.08\textwidth, height=0.08\textwidth]{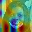}
      \\
    
    \raisebox{4ex}{\makecell{Nose \\ Mouth \\ Exposed}} & \includegraphics[width=0.08\textwidth, height=0.08\textwidth]{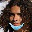}
      & \includegraphics[width=0.08\textwidth, height=0.08\textwidth]{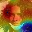}
      &
    \includegraphics[width=0.08\textwidth, height=0.08\textwidth]{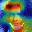}
      &
    \includegraphics[width=0.08\textwidth, height=0.08\textwidth]{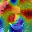}
      \\
\bottomrule
\end{tabular}
      }\hfill
        \subfloat[Chin exposed Grad-CAM]{
        \begin{tabular}{c|cccc}
        \toprule
    \multirow{2}{*}{\textbf{Label}} & \multirow{2}{*}{\textbf{Raw}} & \textbf{BCoP} & \textbf{BCoP} & \multirow{2}{*}{\textbf{FP32}} \\
    & & CNV & $n$-CNV & \\
    \midrule
    \midrule
    \raisebox{4ex}{\makecell{Chin \\ Exposed}} & \includegraphics[width=0.08\textwidth, height=0.08\textwidth]{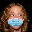}
      & \includegraphics[width=0.08\textwidth, height=0.08\textwidth]{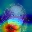}
      &
    \includegraphics[width=0.08\textwidth, height=0.08\textwidth]{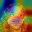}
      &
    \includegraphics[width=0.08\textwidth, height=0.08\textwidth]{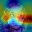}
      \\
    
    \raisebox{4ex}{\makecell{Chin \\ Exposed}} & \includegraphics[width=0.08\textwidth, height=0.08\textwidth]{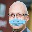}
      & \includegraphics[width=0.08\textwidth, height=0.08\textwidth]{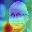}
      &
    \includegraphics[width=0.08\textwidth, height=0.08\textwidth]{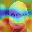}
      &
    \includegraphics[width=0.08\textwidth, height=0.08\textwidth]{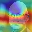}
      \\

    \raisebox{4ex}{\makecell{Chin \\ Exposed}} & \includegraphics[width=0.08\textwidth, height=0.08\textwidth]{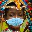}
      & \includegraphics[width=0.08\textwidth, height=0.08\textwidth]{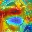}
      &
    \includegraphics[width=0.08\textwidth, height=0.08\textwidth]{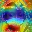}
      &
    \includegraphics[width=0.08\textwidth, height=0.08\textwidth]{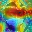}
      \\
    \bottomrule
\end{tabular}}
       \caption{Grad-CAM output of two \bincop{} variants and a float-32 (FP32) CNN. Results are collected for all four wearing positions on a diverse set of individuals. Binarized models show distinct regions of interest which are focused on the exposed part of the face rather than the mask. The FP32 model is difficult to interpret in some cases.}
    \label{fig:classes}
    \vspace{-4ex}
\end{figure*}

\begin{figure}[h]
\centering
\resizebox{\columnwidth}{!}{
\begin{tabular}{c|cccc}
\toprule
\multirow{2}{*}{\textbf{Label}} & \multirow{2}{*}{\textbf{Raw}} & \textbf{BCoP} & \textbf{BCoP} & \multirow{2}{*}{\textbf{FP32}} \\
& & CNV & $n$-CNV & \\
\midrule
\midrule
    \raisebox{3ex}{\makecell{Correctly \\ Masked}} & \includegraphics[width=0.08\textwidth, height=0.08\textwidth]{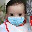}
      & \includegraphics[width=0.08\textwidth, height=0.08\textwidth]{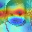}
      &
    \includegraphics[width=0.08\textwidth, height=0.08\textwidth]{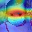}
      &
    \includegraphics[width=0.08\textwidth, height=0.08\textwidth]{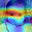}
      \\
    
    \raisebox{3ex}{\makecell{Correctly \\ Masked}} & \includegraphics[width=0.08\textwidth, height=0.08\textwidth]{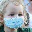}
      & \includegraphics[width=0.08\textwidth, height=0.08\textwidth]{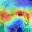}
      &
    \includegraphics[width=0.08\textwidth, height=0.08\textwidth]{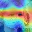}
      &
    \includegraphics[width=0.08\textwidth, height=0.08\textwidth]{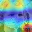}
      \\

    \raisebox{3ex}{\makecell{Correctly \\ Masked}} & \includegraphics[width=0.08\textwidth, height=0.08\textwidth]{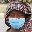}
      & \includegraphics[width=0.08\textwidth, height=0.08\textwidth]{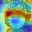}
      &
    \includegraphics[width=0.08\textwidth, height=0.08\textwidth]{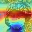}
      &
    \includegraphics[width=0.08\textwidth, height=0.08\textwidth]{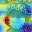}
      \\
      \bottomrule
\end{tabular}}
\caption{Grad-CAM results for age generalization.}
\label{fig:age_gen_lab}
\end{figure}

\begin{figure}[h]
\centering
\resizebox{\columnwidth}{!}{
\begin{tabular}{c|cccc}
\toprule
\multirow{2}{*}{\textbf{Label}} & \multirow{2}{*}{\textbf{Raw}} & \textbf{BCoP} & \textbf{BCoP} & \multirow{2}{*}{\textbf{FP32}} \\
& & CNV & $n$-CNV & \\
\midrule
\midrule
    \raisebox{3ex}{\makecell{Correctly \\ Masked}} & \includegraphics[width=0.08\textwidth, height=0.08\textwidth]{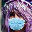}
      & \includegraphics[width=0.08\textwidth, height=0.08\textwidth]{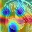}
      &
    \includegraphics[width=0.08\textwidth, height=0.08\textwidth]{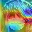}
      &
    \includegraphics[width=0.08\textwidth, height=0.08\textwidth]{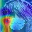}
      \\
    
    \raisebox{3ex}{\makecell{Correctly \\ Masked}} & \includegraphics[width=0.08\textwidth, height=0.08\textwidth]{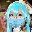}
      & \includegraphics[width=0.08\textwidth, height=0.08\textwidth]{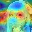}
      &
    \includegraphics[width=0.08\textwidth, height=0.08\textwidth]{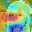}
      &
    \includegraphics[width=0.08\textwidth, height=0.08\textwidth]{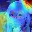}
      \\

    \raisebox{3ex}{\makecell{Nose \\ Exposed}} & \includegraphics[width=0.08\textwidth, height=0.08\textwidth]{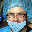}
      & \includegraphics[width=0.08\textwidth, height=0.08\textwidth]{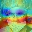}
      &
    \includegraphics[width=0.08\textwidth, height=0.08\textwidth]{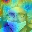}
      &
    \includegraphics[width=0.08\textwidth, height=0.08\textwidth]{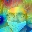}
      \\
  \bottomrule  
\end{tabular}}
\caption{Grad-CAM results for hair/headgear generalization.}
\label{fig:hair_gen}
\end{figure}

\begin{figure}[h]
\centering
\resizebox{\columnwidth}{!}{
\begin{tabular}{c|cccc}
\toprule
\multirow{2}{*}{\textbf{Label}} & \multirow{2}{*}{\textbf{Raw}} & \textbf{BCoP} & \textbf{BCoP} & \multirow{2}{*}{\textbf{FP32}} \\
& & CNV & $n$-CNV & \\
\midrule
\midrule
    \raisebox{3ex}{\makecell{Correctly \\ Masked}} & \includegraphics[width=0.08\textwidth, height=0.08\textwidth]{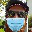}
      & \includegraphics[width=0.08\textwidth, height=0.08\textwidth]{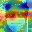}
      &
    \includegraphics[width=0.08\textwidth, height=0.08\textwidth]{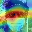}
      &
    \includegraphics[width=0.08\textwidth, height=0.08\textwidth]{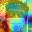}
      \\
    
    \raisebox{3ex}{\makecell{Correctly \\ Masked}} & \includegraphics[width=0.08\textwidth, height=0.08\textwidth]{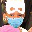}
      & \includegraphics[width=0.08\textwidth, height=0.08\textwidth]{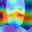}
      &
    \includegraphics[width=0.08\textwidth, height=0.08\textwidth]{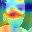}
      &
    \includegraphics[width=0.08\textwidth, height=0.08\textwidth]{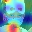}
      \\

    \raisebox{3ex}{\makecell{Chin \\ Exposed}} & \includegraphics[width=0.08\textwidth, height=0.08\textwidth]{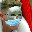}
      & \includegraphics[width=0.08\textwidth, height=0.08\textwidth]{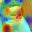}
      &
    \includegraphics[width=0.08\textwidth, height=0.08\textwidth]{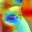}
      &
    \includegraphics[width=0.08\textwidth, height=0.08\textwidth]{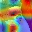}
      \\

    \raisebox{3ex}{\makecell{Nose \\ Mouth \\ Exposed}} & \includegraphics[width=0.08\textwidth, height=0.08\textwidth]{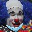}
      & \includegraphics[width=0.08\textwidth, height=0.08\textwidth]{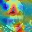}
      &
    \includegraphics[width=0.08\textwidth, height=0.08\textwidth]{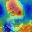}
      &
    \includegraphics[width=0.08\textwidth, height=0.08\textwidth]{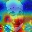}
      \\
    
    \raisebox{3ex}{\makecell{Nose \\ Mouth \\ Exposed}} & \includegraphics[width=0.08\textwidth, height=0.08\textwidth]{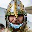}
      & \includegraphics[width=0.08\textwidth, height=0.08\textwidth]{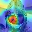}
      &
    \includegraphics[width=0.08\textwidth, height=0.08\textwidth]{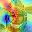}
      &
    \includegraphics[width=0.08\textwidth, height=0.08\textwidth]{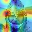}
      \\
      \bottomrule
\end{tabular}}
\caption{Grad-CAM results for face manipulation with double-masks, face paint and sunglasses.}
\label{fig:face_manipulation}
\end{figure}

We further analyze the output heat maps generated by Grad-CAM to interpret the predictions of our BNNs with respect to the diverse attributes of the MaskedFace-Net dataset. In Fig.~\ref{fig:classes} and Fig.~\ref{fig:age_gen_lab} - Fig.~\ref{fig:face_manipulation}, column 1 and 2 indicate the label and input image respectively. Columns 3, 4 and 5 highlight the heat maps obtained from the Grad-CAM output of \bincop{}-CNV, \bincop{}-$n$-CNV and a full-precision version of CNV with float-32 parameters (FP32). The heat maps are overlaid on the raw input images for better visualization. All raw images chosen have been classified correctly by all the networks, for fair interpretation of feature-to-prediction correlation.

%Fig 2
In Fig.~\ref{fig:classes}(a), we analyze the Region of Interest (RoI) for the \textit{correctly masked} class. \bincop{}'s learning capacity allows it to focus on key facial lineaments of the human wearing the mask, rather than the mask itself. This potentially helps in generalizing on other mask types. For the child example shown in the first row, the focus of \bincop{} lies on the nose area, asserting that it is fully covered to result in a correctly masked prediction. Similarly, for the adult in row 2, \bincop{}-CNV focuses on the upper edge of the mask, to predict its coverage of the face. This also holds for our small version of \bincop{}, with significantly reduced learning capacity. The RoI curves finely above the mask, tracing the exposed region of the face. In the third row example, \bincop{}-CNV falls back to focusing on the mask, whereas \bincop{}-$n$-CNV continues to focus on the exposed features. Both models achieve the same prediction by focusing on different parts of the raw image. In contrast to the \bincop{} variants, the full-precision FP32 model seems to focus on a combination of several different features on all three examples. This can be attributed to its larger learning capacity and possible overfitting.

In Fig.~\ref{fig:classes}(b), we analyze the Grad-CAM output of the \textit{uncovered nose} class. \bincop{}-CNV and \bincop{}-$n$-CNV focus specifically on two regions, namely the nose and the straight upper edge of the mask. These clear characteristics cannot be observed with the oversized FP32 CNN. In Fig~\ref{fig:classes}(c), the results show the RoI for predicting the \textit{exposed mouth and nose} class. All models seem to distribute their attention onto several exposed features of the face. Fig.~\ref{fig:classes}(d) shows Grad-CAM results for \textit{chin exposed} predictions. Although the top region of the mask points upwards, similar to the correctly worn mask, the BNNs pay less attention to this region and instead focus on the neck and chin. With the full-precision FP32 model, it is difficult to interpret the reason for the correct classification, as little to no focus is given to the chin region, again hinting at possible overfitting.

Beyond studying the BNNs' behavior on different class predictions, we can use the attention heat maps to understand the generalization behavior of the classifier. In Fig.~\ref{fig:age_gen_lab} - Fig.~\ref{fig:face_manipulation}, we test \bincop{}'s generalization over ages, hair colors and head gear, as well as complete face manipulation with double-masks, face paint and sunglasses. In Fig~\ref{fig:age_gen_lab}, we see that the smaller eyes of infants and elderly do not hinder \bincop{}'s ability to focus on the top region of the correctly worn masks. In Fig.~\ref{fig:hair_gen}, \bincop{}-CNV shows resilience to differently colored hair and head-gear, even when having a similar light-blue color as the face-masks (row 2 and 3). In contrast, the FP32 model's attention seems to shift towards the hair and head-gear for these cases. Finally, in Fig.~\ref{fig:face_manipulation}, both \bincop{} variants focus on relevant features of the corresponding label, irrespective of the obscured or manipulated faces. This empirically shows that the complex training of BNNs, along with their lower information capacity, constrains them to focus on a smaller set of relevant features, thereby generalizing well for unprecedented cases.

\subsection{Discussion and Comparison with Other Works}
As mentioned in Sec~\ref{sec:related_covid}, detection of masks has piqued the interest of many researchers in the computer vision domain due to its relevance in the context of the ongoing COVID-19 pandemic. NVIDIA proposed mask recognition using object detection models~\cite{nvidia_covid}. These models require INT8 or Float-16 numerical precision, with ResNet-18 as a backbone for input images of 960$\times$544. The complexity is orders of magnitude higher than the models we propose in this paper. A head-to-head comparison is difficult to make due to differences in the training approach, the CNN model, the datasets used and the application requirements. The networks are trained to predict only two classes (mask, no mask), which is a simpler problem compared to the exact positioning supported by \bincop{}. However, the localization and higher resolution makes it a more complex task overall. With the NVIDIA Jetson Nano hardware, which typically requires $\sim$10W of power on intensive workloads, a frame-rate of 21 FPS is achieved. The more powerful 25W Jetson AGX Xavier can achieve up to 508 FPS. Compared to the NVIDIA approach~\cite{nvidia_covid}, \bincop{} is targeted at low-power, embedded applications with peak inference power of $\sim$2W and high classification rates of up to $\sim$6400 FPS on smaller resolution input images. It is worth noting that \bincop{} can also classify high resolution images containing multiple individuals, by slicing the input into many 32$\times$32 frames and batch processing them. This application makes use of the high-throughput results presented in Tab.~\ref{tab:dse_results}. Another approach proposed by Agarwal et al.~\cite{amazonmask} achieves the task of detecting a range of personal protective equipment (PPE). Processing takes place on cloud servers, which could raise privacy and data safety concerns in public settings. Wang et al.~\cite{wang2021wearmask} propose an in-browser server-less edge computing method, with object detection models. The browser-enabled device must support the WebAssembly instruction format. The authors benchmarked their approach on an iPad Pro (A9X), an iPhone 11 (A13) and a MacBook pro (Intel i7-9750H), achieving 5, 10 and 20 FPS respectively. Needless to say, these devices (or similar) are expensive and cannot be placed in abundance in public areas. Similarly,~\cite{cym} offers an Android application solution, which is suitable for users self-checking their masks. In this case, low-power, edge-hardware and continuous surveillance are not emphasized.

Our approach offers a unique, low-power, high-throughput solution, which is applicable to cheap, embedded FPGAs. Moreover, the \bincop{} solution is not constrained to FPGA platforms. Software-based inference of~\bincop{} is also possible on other low-power microcontrollers, with binary instructions. Training on synthetic data allows us to generate more samples with different mask colors, shapes, and sizes~\cite{anwar2020masked}, further improving the generalizability of the BNNs, while keeping real-world data available for fine-tuning stages. %Optionally, a face detector stage can be added before \bincop{} to filter out images with no faces and further enhance the crowd classification scenario.
%Synthetic data can also be combined with real-world datasets, which are difficult to collect and only available in limited sizes.
%%%%%%%%%%%%%%%%%%%%%%%%%%%%%%%%%%%%%%%%%%%%%%%%%%%%%%%%%%%%%%%%%%%%%%%%%%%%%%%%
% __   __        __             __     __       
%/  ` /  \ |\ | /  ` |    |  | /__` | /  \ |\ | 
%\__, \__/ | \| \__, |___ \__/ .__/ | \__/ | \| 
%%%%%%%%%%%%%%%%%%%%%%%%%%%%%%%%%%%%%%%%%%%%%%%%%%%%%%%%%%%%%%%%%%%%%%%%%%%%%%%%
\section{Conclusion}
In this paper, we apply binary neural networks to the task of classifying the correctness of face-mask wear and positioning. In the context of the ongoing COVID-19 pandemic, such algorithms can be used at entrances to corporate buildings, airports, shopping areas, and other indoor locations to mitigate the spread of the virus. Applying BNNs to this application solves several challenges such as (1) maintaining data privacy of the public by processing data on the edge-device, (2) deploying the classifier on an efficient XNOR-based accelerator to achieve low-power computation, and (3) minimizing the neural network's memory footprint by representing all parameters in the binary domain, enabling deployment on low-cost, embedded hardware. The accelerator requires only $\sim$1.65W of power when idling on single gates/entrances. Alternatively, high-performance is possible, providing fast batch classification on multiple gates and entrances with multiple cameras, at $\sim$6400 frames-per-second and 2W of power. We achieve an accuracy of up to 98\% for four wearing positions of the MaskedFace-Net dataset. The Grad-CAM approach is used to study the features learned by the proposed \bincop{} classifier. The results show the classifier's high generalization ability, allowing it to perform well on different face structures, skin-tones, hair types, and age groups.

\bibliographystyle{IEEEtran}
\bibliography{IEEEbib.bib}

\begin{thebibliography}{10}
\providecommand{\url}[1]{#1}
\csname url@rmstyle\endcsname
\providecommand{\newblock}{\relax}
\providecommand{\bibinfo}[2]{#2}
\providecommand\BIBentrySTDinterwordspacing{\spaceskip=0pt\relax}
\providecommand\BIBentryALTinterwordstretchfactor{4}
\providecommand\BIBentryALTinterwordspacing{\spaceskip=\fontdimen2\font plus
\BIBentryALTinterwordstretchfactor\fontdimen3\font minus
  \fontdimen4\font\relax}
\providecommand\BIBforeignlanguage[2]{{%
\expandafter\ifx\csname l@#1\endcsname\relax
\typeout{** WARNING: IEEEtran.bst: No hyphenation pattern has been}%
\typeout{** loaded for the language `#1'. Using the pattern for}%
\typeout{** the default language instead.}%
\else
\language=\csname l@#1\endcsname
\fi
#2}}

\bibitem{LeNet}
Y.~{Lecun}, L.~{Bottou}, Y.~{Bengio}, and P.~{Haffner}, ``Gradient-based
  learning applied to document recognition,'' \emph{Proceedings of the IEEE},
  vol.~86, no.~11, pp. 2278--2324, 1998.

\bibitem{covidnet}
L.~Wang, Z.~Q. Lin, and A.~Wong, ``Covid-net: A tailored deep convolutional
  neural network design for detection of covid-19 cases from chest x-ray
  images,'' \emph{Scientific Reports}, vol.~10, no.~1, pp. 1--12, 2020.

\bibitem{CoroNet}
\BIBentryALTinterwordspacing
A.~I. Khan, J.~L. Shah, and M.~M. Bhat, ``Coronet: A deep neural network for
  detection and diagnosis of covid-19 from chest x-ray images,'' \emph{Computer
  Methods and Programs in Biomedicine}, vol. 196, p. 105581, Nov 2020.
  [Online]. Available: \url{http://dx.doi.org/10.1016/j.cmpb.2020.105581}
\BIBentrySTDinterwordspacing

\bibitem{WHO}
\BIBentryALTinterwordspacing
``When and how to use masks.'' [Online]. Available:
  \url{https://www.who.int/emergencies/diseases/novel-coronavirus-2019/advice-for-public/when-and-how-to-use-masks}
\BIBentrySTDinterwordspacing

\bibitem{tradCVvsAI}
N.~O'Mahony, S.~Campbell, A.~Carvalho, S.~Harapanahalli, G.~V. Hernandez,
  L.~Krpalkova, D.~Riordan, and J.~Walsh, ``Deep learning vs. traditional
  computer vision,'' in \emph{Advances in Computer Vision}, K.~Arai and
  S.~Kapoor, Eds.\hskip 1em plus 0.5em minus 0.4em\relax Cham: Springer
  International Publishing, 2020, pp. 128--144.

\bibitem{maskedfacenet}
\BIBentryALTinterwordspacing
A.~Cabani, K.~Hammoudi, H.~Benhabiles, and M.~Melkemi, ``Maskedface-net -- a
  dataset of correctly/incorrectly masked face images in the context of
  covid-19,'' \emph{Smart Health}, 2020. [Online]. Available:
  \url{http://www.sciencedirect.com/science/article/pii/S2352648320300362}
\BIBentrySTDinterwordspacing

\bibitem{FINN}
\BIBentryALTinterwordspacing
Y.~Umuroglu, N.~J. Fraser, G.~Gambardella, M.~Blott, P.~Leong, M.~Jahre, and
  K.~Vissers, ``Finn: A framework for fast, scalable binarized neural network
  inference,'' in \emph{Proceedings of the 2017 ACM/SIGDA International
  Symposium on Field-Programmable Gate Arrays}, ser. FPGA '17.\hskip 1em plus
  0.5em minus 0.4em\relax New York, NY, USA: ACM, 2017, pp. 65--74. [Online].
  Available: \url{http://doi.acm.org/10.1145/3020078.3021744}
\BIBentrySTDinterwordspacing

\bibitem{FaceMaskGer}
\BIBentryALTinterwordspacing
T.~Mitze, R.~Kosfeld, J.~Rode, and K.~W{\"a}lde, ``Face masks considerably
  reduce covid-19 cases in germany,'' \emph{Proceedings of the National Academy
  of Sciences}, vol. 117, no.~51, pp. 32\,293--32\,301, 2020. [Online].
  Available: \url{https://www.pnas.org/content/117/51/32293}
\BIBentrySTDinterwordspacing

\bibitem{MAFA}
S.~{Ge}, J.~{Li}, Q.~{Ye}, and Z.~{Luo}, ``Detecting masked faces in the wild
  with lle-cnns,'' in \emph{2017 IEEE Conference on Computer Vision and Pattern
  Recognition (CVPR)}, 2017, pp. 426--434.

\bibitem{wang2020masked}
Z.~{Wang}, G.~{Wang}, B.~{Huang}, Z.~{Xiong}, Q.~{Hong}, H.~{Wu}, P.~{Yi},
  K.~{Jiang}, N.~{Wang}, Y.~{Pei}, H.~{Chen}, Y.~{Miao}, Z.~{Huang}, and
  J.~{Liang}, ``{Masked Face Recognition Dataset and Application},''
  \emph{arXiv e-prints}, p. arXiv:2003.09093, Mar. 2020.

\bibitem{BNN}
\BIBentryALTinterwordspacing
I.~Hubara, M.~Courbariaux, D.~Soudry, R.~El-Yaniv, and Y.~Bengio, ``Binarized
  neural networks,'' in \emph{Advances in Neural Information Processing Systems
  29}.\hskip 1em plus 0.5em minus 0.4em\relax Curran Associates, Inc., 2016,
  pp. 4107--4115. [Online]. Available:
  \url{http://papers.nips.cc/paper/6573-binarized-neural-networks.pdf}
\BIBentrySTDinterwordspacing

\bibitem{XNOR}
M.~Rastegari, V.~Ordonez, J.~Redmon, and A.~Farhadi, ``{XNOR-Net: ImageNet
  Classification Using Binary Convolutional Neural Networks},'' in \emph{The
  European Conference on Computer Vision (ECCV)}.\hskip 1em plus 0.5em minus
  0.4em\relax Cham: Springer International Publishing, 2016, pp. 525--542.

\bibitem{BinaryConnect}
M.~Courbariaux, Y.~Bengio, and J.-P. David, ``Binaryconnect: Training deep
  neural networks with binary weights during propagations,'' in \emph{Advances
  in Neural Information Processing Systems (NeurIPS)}, C.~Cortes, N.~D.
  Lawrence, D.~D. Lee, M.~Sugiyama, and R.~Garnett, Eds.\hskip 1em plus 0.5em
  minus 0.4em\relax Curran Associates, Inc., 2015, pp. 3123--3131.

\bibitem{ABCNET}
\BIBentryALTinterwordspacing
X.~Lin, C.~Zhao, and W.~Pan, ``Towards accurate binary convolutional neural
  network,'' in \emph{Advances in Neural Information Processing Systems 30},
  I.~Guyon, U.~V. Luxburg, S.~Bengio, H.~Wallach, R.~Fergus, S.~Vishwanathan,
  and R.~Garnett, Eds.\hskip 1em plus 0.5em minus 0.4em\relax Curran
  Associates, Inc., 2017, pp. 345--353. [Online]. Available:
  \url{http://papers.nips.cc/paper/6638-towards-accurate-binary-convolutional-neural-network.pdf}
\BIBentrySTDinterwordspacing

\bibitem{bdad_fri_20b}
A.~{Frickenstein}, M.~{Rohit Vemparala}, J.~{Mayr}, N.~{Shankar Nagaraja},
  C.~{Unger}, F.~{Tombari}, and W.~{Stechele}, ``{Binary DAD-Net: Binarized
  Driveable Area Detection Network for Autonomous Driving},'' \emph{arXiv
  e-prints}, p. arXiv:2006.08178, June 2020.

\bibitem{7878541}
R.~{Andri}, L.~{Cavigelli}, D.~{Rossi}, and L.~{Benini}, ``Yodann: An
  architecture for ultralow power binary-weight cnn acceleration,'' \emph{IEEE
  Transactions on Computer-Aided Design of Integrated Circuits and Systems},
  vol.~37, no.~1, pp. 48--60, Jan 2018.

\bibitem{BRein}
K.~{Ando}, K.~{Ueyoshi}, K.~{Orimo}, H.~{Yonekawa}, S.~{Sato}, H.~{Nakahara},
  S.~{Takamaeda-Yamazaki}, M.~{Ikebe}, T.~{Asai}, T.~{Kuroda}, and
  M.~{Motomura}, ``Brein memory: A single-chip binary/ternary reconfigurable
  in-memory deep neural network accelerator achieving 1.4 tops at 0.6 w,''
  \emph{IEEE Journal of Solid-State Circuits}, vol.~53, no.~4, pp. 983--994,
  April 2018.

\bibitem{towardshwBNN}
\BIBentryALTinterwordspacing
C.~Fu, S.~Zhu, H.~Su, C.-E. Lee, and J.~Zhao, ``Towards fast and
  energy-efficient binarized neural network inference on fpga,'' in
  \emph{Proceedings of the 2019 ACM/SIGDA International Symposium on
  Field-Programmable Gate Arrays}, ser. FPGA '19.\hskip 1em plus 0.5em minus
  0.4em\relax New York, NY, USA: Association for Computing Machinery, 2019, p.
  306. [Online]. Available: \url{https://doi.org/10.1145/3289602.3293990}
\BIBentrySTDinterwordspacing

\bibitem{STE}
\BIBentryALTinterwordspacing
Y.~Bengio, N.~L{\'{e}}onard, and A.~C. Courville, ``Estimating or propagating
  gradients through stochastic neurons for conditional computation,''
  \emph{CoRR}, vol. abs/1308.3432, 2013. [Online]. Available:
  \url{http://arxiv.org/abs/1308.3432}
\BIBentrySTDinterwordspacing

\bibitem{cam}
B.~{Zhou}, A.~{Khosla}, A.~{Lapedriza}, A.~{Oliva}, and A.~{Torralba},
  ``Learning deep features for discriminative localization,'' in \emph{IEEE/CVF
  Conference on Computer Vision and Pattern Recognition (CVPR)}, June 2016, pp.
  2921--2929.

\bibitem{Selvaraju_2017_ICCV}
R.~R. Selvaraju, M.~Cogswell, A.~Das, R.~Vedantam, D.~Parikh, and D.~Batra,
  ``Grad-cam: Visual explanations from deep networks via gradient-based
  localization,'' in \emph{Proceedings of the IEEE International Conference on
  Computer Vision (ICCV)}, Oct 2017.

\bibitem{cifar10}
A.~Krizhevsky, ``Learning multiple layers of features from tiny images,''
  \emph{University of Toronto}, 2009.

\bibitem{orthruspe}
N.~{Fasfous}, M.~R. {Vemparala}, A.~{Frickenstein}, and W.~{Stechele},
  ``Orthruspe: Runtime reconfigurable processing elements for binary neural
  networks,'' in \emph{2020 Design, Automation Test in Europe Conference
  Exhibition (DATE)}, 2020, pp. 1662--1667.

\bibitem{VGG}
K.~Simonyan and A.~Zisserman, ``Very deep convolutional networks for
  large-scale image recognition,'' in \emph{International Conference on
  Learning Representations}, 2015.

\bibitem{nvidia_covid}
\BIBentryALTinterwordspacing
A.~Kulkarni, A.~Vishwanath, and C.~Shah, ``Implementing a real-time, ai-based,
  face mask detector application for covid-19,'' Feb 2021. [Online]. Available:
  \url{https://developer.nvidia.com/blog/implementing-a-real-time-ai-based-face-mask-detector-application-for-covid-19/}
\BIBentrySTDinterwordspacing

\bibitem{amazonmask}
\BIBentryALTinterwordspacing
T.~Agrawal, K.~Imran, M.~Figus, and C.~Kirkpatrick, ``Automatically detecting
  personal protective equipment on persons in images using amazon
  rekognition,'' Oct 2020. [Online]. Available:
  \url{https://aws.amazon.com/blogs/machine-learning/automatically-detecting-personal-protective-equipment-on-persons-in-images-using-amazon-rekognition/}
\BIBentrySTDinterwordspacing

\bibitem{wang2021wearmask}
Z.~{Wang}, P.~{Wang}, P.~C. {Louis}, L.~E. {Wheless}, and Y.~{Huo},
  ``{WearMask: Fast In-browser Face Mask Detection with Serverless Edge
  Computing for COVID-19},'' \emph{arXiv e-prints}, p. arXiv:2101.00784, Jan.
  2021.

\bibitem{cym}
\BIBentryALTinterwordspacing
K.~Hammoudi, A.~Cabani, H.~Benhabiles, and M.~Melkemi, ``Validating the correct
  wearing of protection mask by taking a selfie: Design of a mobile application
  “checkyourmask” to limit the spread of covid-19,'' \emph{Computer
  Modeling in Engineering \& Sciences}, vol. 124, no.~3, pp. 1049--1059, 2020.
  [Online]. Available: \url{http://www.techscience.com/CMES/v124n3/39927}
\BIBentrySTDinterwordspacing

\bibitem{anwar2020masked}
A.~{Anwar} and A.~{Raychowdhury}, ``{Masked Face Recognition for Secure
  Authentication},'' \emph{arXiv e-prints}, p. arXiv:2008.11104, Aug. 2020.

\end{thebibliography}

\end{document}